\documentclass[letterpaper]{article} 
\usepackage{aaai25}  
\usepackage{times}  
\usepackage{helvet}  
\usepackage{courier}  
\usepackage[hyphens]{url}  
\usepackage{graphicx} 
\usepackage{natbib}  
\usepackage{caption} 
\usepackage{algorithm}
\usepackage[noEnd]{algpseudocodex}

\algrenewcommand\algorithmicindent{1.0em}

\usepackage{caption}
\usepackage{subcaption}
\usepackage{amsfonts}
\usepackage{amsmath}
\usepackage{booktabs}
\usepackage{multirow}
\usepackage{tabularx}
\usepackage[T1]{fontenc}
\usepackage{float}
\newcommand\thefont{\expandafter\string\the\font}

\usepackage{newfloat}
\usepackage{listings}
\DeclareCaptionStyle{ruled}{labelfont=normalfont,labelsep=colon,strut=off} 
\floatstyle{ruled}
\newfloat{listing}{tb}{lst}{}
\floatname{listing}{Listing}

\title{Neural Combinatorial Optimization for Stochastic Flexible Job Shop Scheduling Problems
}

\title{Neural Combinatorial Optimization for Stochastic Flexible Job Shop Scheduling Problems}
\author {
    Igor G. Smit\textsuperscript{\rm 1, \rm 2},
    Yaoxin Wu\textsuperscript{\rm 2, \rm 3}\thanks{Corresponding author.},
    Pavel Troubil\textsuperscript{\rm 4},
    Yingqian Zhang\textsuperscript{\rm 2, \rm 3},
    Wim P.M. Nuijten\textsuperscript{\rm 1, \rm 2}
}
\affiliations {
    \textsuperscript{\rm 1}Department of Mathematics and Computer Science, Eindhoven University of Technology\\
    \textsuperscript{\rm 2}Eindhoven Artificial Intelligence Systems Institute, Eindhoven University of Technology\\
    \textsuperscript{\rm 3}Department of Industrial Engineering and Innovation Sciences, Eindhoven University of Technology\\
    \textsuperscript{\rm 4}Delmia R\&D, Dassault Systèmes\\
    i.g.smit@tue.nl, y.wu2@tue.nl, pavel.troubil@3ds.com, yqzhang@tue.nl, w.p.m.nuijten@tue.nl
}

\begin{document}

\maketitle

\begin{abstract}
Neural combinatorial optimization (NCO) has gained significant attention due to the potential of deep learning to efficiently solve combinatorial optimization problems. NCO has been widely applied to job shop scheduling problems (JSPs) with the current focus predominantly on deterministic problems. In this paper, we propose a novel attention-based scenario processing module (SPM) to extend NCO methods for solving stochastic JSPs. Our approach explicitly incorporates stochastic information by an attention mechanism that captures the embedding of sampled scenarios (i.e., an approximation of stochasticity). Fed with the embedding, the base neural network is intervened by the attended scenarios, which accordingly learns an effective policy under stochasticity. We also propose a training paradigm that works harmoniously with either the expected makespan or Value-at-Risk objective. Results demonstrate that our approach outperforms existing learning and non-learning methods for the flexible JSP problem with stochastic processing times on a variety of instances. In addition, our approach holds significant generalizability to varied numbers of scenarios and disparate distributions.
\end{abstract}

\begin{links}
    \link{Code}{https://github.com/ai-for-decision-making-tue/NCO-for-Stochastic-FJSP}
\end{links}

\section{Introduction}
Combinatorial optimization (CO) problems are prevalent in various application areas, posing significant challenges due to their complex nature and computational intractability. One critical CO problem is the flexible job shop scheduling problem (FJSP), which has wide applicability ranging from semiconductor manufacturing \cite{Tamssaouet2022} to healthcare scheduling \cite{Burdett2018} and aluminum production \cite{Zhang2016}. Traditional approaches, such as constraint programming \cite{Baptiste2001,Col2022}, heuristics \cite{Sels2012}, or meta-heuristics \cite{Rooyani2019} have made great progress in solving these problems. However, these methods generally assume deterministic problems. In practice, oppositely, there is an abundance of uncertainty, leading to vulnerable plans produced by these optimization methods. 
\par
In recent years, increasing focus has been directed to stochastic optimization methods. However, these methods remain largely underrepresented, especially for more complex optimization problems such as the FJSP. Existing methods solving the stochastic FJSP, such as a simulation-optimization framework \cite{Ghaedy2024} or a meta-heuristic approach including Monte-Carlo sampling \cite{Flores2023} add significant computational complexity to already expensive approaches, while also being tailored to specific use-cases and objectives, limiting their potential for adoption.  
\par
A recent branch of optimization technology is constituted by neural combinatorial optimization (NCO) methods, leveraging deep (reinforcement) learning (DRL) to solve CO problems~\cite{bengio2021machine}. They learn high-quality optimization policies that fit a set of problem instances, eliminating the need for handcrafted expert rules, while also being efficient. 
Current NCO methods mainly target deterministic optimization, overlooking their practical use in stochastic situations. It is highly beneficial to extend NCO to the stochastic optimization domain.
\par
Several existing works leverage learning-based methods for stochastic routing or scheduling. However, these are generally focused on online or dynamic scenarios. Conversely, in many practical cases, such as manufacturing plants, a one-time schedule is needed in advance. Moreover, existing methods only focus on implicitly learning stochastic dynamics through the Markov decision process (MDP). In stochastic optimization, on the contrary, the assumed distributions can be explicitly considered, allowing for better plans.
\par
We address these issues by a novel attention-based \textit{scenario processing module} (SPM) that extends deterministic NCO architectures to solve stochastic problems. The SPM, which is expressive yet easily adaptable to different base neural networks for different problems, effectively captures the representation of stochastic scenarios to favorably intervene the solution policy. We embed SPM in a novel training paradigm that can address various stochastic objectives and offers high-quality yet fast solution policies, using a sampling strategy commonly found in stochastic optimization methods. We combine SPM with an existing network architecture to form SPM-DAN and solve the stochastic FJSP. 
Results show our method outperforms existing methods consistently. Our main contributions are summarized as follows:
\begin{enumerate}
    \item We propose an attention-based SPM to solve stochastic optimization problems, which is expressive and modular, with a good transferability to different base neural networks.
    \item We put forward an effective DRL training paradigm, which is used to optimize different stochastic objectives.
    \item We apply SPM to solve the stochastic FJSP, which outperforms existing learning and non-learning methods.
\end{enumerate}

\section{Related Work}
\paragraph{NCO for FJSP} Graph neural networks (GNNs) for solving a variety of scheduling problems have been developed in recent years \cite{zhang2020, Park2021, Lei2022, Kwon2021}. We refer to a recent survey \cite{smit2025} for a complete outline. \citet{song2022flexible} first proposed a competitive end-to-end DRL algorithm to construct FJSP solutions. They used a heterogeneous graph and designed a heterogeneous GNN using different GAT \cite{veličković2018graphattentionnetworks} layers to encode machine and operation nodes. In the follow-up work, \citet{wang2023flexible} proposed a recent state-of-the-art dual attention network (DAN) that comprises both self- and cross-attention, which achieved superior performance over previous DRL approaches.

\paragraph{NCO for Dynamic FJSP} Multiple papers target the dynamic FJSP \cite[e.g.][]{zhang2023deep, lei2023large}, which involve different sources of uncertainty such as random arrivals, machine availability, or processing times. These papers mainly focus on developing training strategies and MDP formulations to fit the dynamic use cases. The network architectures are not specifically tailored to address uncertainties, which are implicitly handled by approximating the expected reward in MDP trajectories. 
Similarly, NCO methods for other problems also mainly target deterministic or dynamic cases, in which the realized values become known during solution construction \cite{Schmitt2022,Kwon2020,Joe_Lau_2020}. 

\paragraph{NCO for Stochastic FJSP} Similar to us, \citet{Infantes2024} consider a stochastic job shop scheduling problem (JSP) for which a full plan must be created before the realizations become known. They include three features in the state space to describe the assumed triangular distribution, and base the reward function on a single sampled realization for each instance to optimize the expected makespan. While promising, their method is limited to triangular processing time distributions and the expected makespan objective.   

\section{Preliminaries and Problem Formulation}
\paragraph{Flexible Job Shop Scheduling Problem}
The FJSP consists of a pair $(\mathcal{J}, \mathcal{M})$ where $\mathcal{J}$ and $\mathcal{M}$ are jobs and  machines, respectively. A job $J_i \in \mathcal{J}$ consists of a set $\mathcal{O}_i = \{O_{i1}, \dots, O_{in_{i}}\}$ of $n_i$ operations to be performed in order. The total set of operations is $\mathcal{O} = \bigcup_i \mathcal{O}_i$. Each operation $O_{ij}$ must be processed by a single machine, selected from the set of compatible machines $\mathcal{M}_{ij} \subseteq \mathcal{M}$. The processing time to perform operation $O_{ij}$ on machine $M_k \in \mathcal{M}_{ij}$ is given by $p_{ij}^k > 0$ and each machine can only process one job at a time. A solution to the FJSP is a \textit{schedule}, which assigns a compatible machine to each operation $O_{ij}$ and determines the order of operations on each machine. The goal is to minimize the makespan $c_{max} = \max_{O_{ij} \in \mathcal{O}}c_{ij}$, which is the maximum completion time $c_{ij}$ of all operations.

\paragraph{Stochastic Flexible Job Shop Scheduling Problem}
In the stochastic FJSP, we assume that the processing times are random variables $P_{ij}^k$. The realized processing times $p_{ij}^k \sim {\mathcal{P}_{ij}^k}$ follow the probability distributions $\mathcal{P}_{ij}^k$ and only become known after the schedule is created. As a result, the realized $c_{ij}$ and $c_{max}$ are also realizations of the random variables $C_{ij}$ and $C_{max}$. The goal is to create a schedule that optimizes an objective function $f(C_{max})$, which can be the expected value $\mathbb{E}\left(C_{max}\right)$, the Value-at-Risk $VaR_\alpha\left({C_{max}}\right)=\min\{c_{max}: \mathbb{P}(C_{max} \leq c_{max}) \geq \alpha\}$, or other functions of $C_{max}$. $VaR_\alpha$ indicates the value for which, with probability $\alpha$, our makespan is at most this high. As such, it considers the practically relevant robustness to the uncertainty of schedules. We set and keep $\alpha=95\%$ in this paper.  

\paragraph{Attention Mechanism}
The attention function aims to map a group of $n$ queries $Q \in \mathbb{R}^{n\times d_q}$ to their advanced representations, using $n_v$ keys $K \in \mathbb{R}^{n_v\times d_q}$ and $n_v$ values $V\in \mathbb{R}^{n_v\times d_v}$, such that:
{
\[\text{Att}(Q,K,V) = \text{softmax}(QK^T/\sqrt{d_q})V\]}

\noindent where $QK^T$ defines the similarity between $Q$ and $K$, based on which weights are computed by the softmax function and then used to obtain a weighted sum of the values in $V$.
\par
\citet{Vaswani2017} extended the attention mechanism by introducing the multi-head attention. By setting $h$ heads, $Q$, $K$, and $V$ are first projected to $h$ matrices, respectively. Then, the attention function is applied to every three matrices in each head. The advanced representations from all heads are concatenated and linearly transformed:
{
\[\text{MHA}(Q, K, V) = \text{concat}(\text{head}_1, \dots, \text{head}_h)W^O\]}

\noindent where $\text{head}_i = \text{Att}(QW_i^Q,KW_i^k,VW_i^V)$. 
$W_i^Q \in \mathbb{R}^{d_q\times d_q'}$, $W_i^K \in \mathbb{R}^{d_q\times d_q'}$, and $W_i^V \in \mathbb{R}^{d_v\times d_v'}$ are learnable matrices. We follow a typical choice for the dimensions by setting $d_q=d_v=d$ and $d_q'=d_v'=d/h$. The multi-head attention layer is often embedded in a block, alongside the normalization, skip-connection, and feed-forward layer. 
Given input matrices $X\in\mathbb{R}^{n\times d}$ and $Y\in\mathbb{R}^{n_v\times d}$ a~multi-head attention block is defined as:
{
\[\text{MHAB}(X,Y) = \text{LayerNorm}(Z + \text{FF}(Z))\]}
\noindent where $Z = \text{LayerNorm}(X + \text{MHA}(X,Y,Y)$. 
$\text{FF}$ is a fully-connected feed-forward layer and \text{LayerNorm} is the layer normalization operation \cite{Ba2016}.

\section{Methodology} 
We take the state-of-the-art FJSP method, i.e., the DAN from \cite{wang2023flexible}, for an example, and describe how our method SPM-DAN is developed from DAN. Our method is easily applicable to other NCO methods.

\subsection{Markov Decision Process}
The scheduling process can be considered as a sequential decision-making process of iteratively assigning operations to available machines. At every decision moment $t$, an operation-machine pair $(O_{ij}, M_k)$ is selected such that $O_{ij}$ can be assigned to $M_k$. In the MDP formulation, the agent at each step receives state $s_t$, representing the environment, and takes action $a_t = (O_{ij}, M_k) \in \mathcal{A}(t)$ from the set of eligible actions $\mathcal{A}(t)$, which are the possible assignments of the first unscheduled operation in each job to a compatible machine. The environment feeds back reward $r_t$ and new state $s_{t+1}$. The schedule is completed after $|\mathcal{O}|$ actions.

\subsubsection{State Space}
The relevant operations $\mathcal{O}_u(t) \subseteq \mathcal{O}$ for the state $s_t$ are all operations except those that already have a successor scheduled on the same machine and, thus, do not directly influence the schedule anymore. The relevant machines $\mathcal{M}_u(t) \subseteq \mathcal{M}$ are all machines on which any of the remaining operations can be scheduled. For the deterministic method proposed in \cite{wang2023flexible}, the state $s_t = \{\mathcal{H}_O, \mathcal{H}_M, \mathcal{H}_{OM}\}$ consists of the operation features $\mathcal{H}_O = \{h_{O_{ij}} \in \mathbb{R}^{10} | O_{ij} \in \mathcal{O}_u(t)\}$, machine features $\mathcal{H}_M = \{h_{M_k} \in \mathbb{R}^8 | M_k \in \mathcal{M}_u(t)\}$, and operation-machine pair features $\mathcal{H}_{OM} = \{h_{(O_{ij}, M_k)} \in \mathbb{R}^8 | (O_{ij}, M_k) \in \mathcal{A}(t)\}$. These features are dynamic and interdependent.
We refer to the original paper for the detailed feature descriptions. We introduce an expansion of this state space, based on sample approximation. Concretely, we sample $n_{scn}$ independent scenarios, each representing a set of realized processing time values $p_{ij}^k \sim {\mathcal{P}_{ij}^k}$. For each scenario, we keep track of all the features during the scheduling process. In addition, we maintain the deterministic problem instance. In doing so, the expanded state captures both the deterministic and stochastic problem characteristics. Formally, our state at step $t$ is $s_t = \{s_t^{det}, \mathcal{S}_t^{stoch}\}$, where $s_t^{det} = \{\mathcal{H}_O^{det}, \mathcal{H}_M^{det}, \mathcal{H}_{OM}^{det}\}$ represents the deterministic instance state, and $\mathcal{S}_t^{stoch} = \{s_t^l | 1 \leq l \leq n_{scn}\}$ is the set of states representing the scenarios $s_t^l =  \{\mathcal{H}_O^{l}, \mathcal{H}_M^{l}, \mathcal{H}_{OM}^{l}\}$. Note that the sets $\mathcal{O}_u(t)$, $\mathcal{M}_u(t)$, and $\mathcal{A}(t)$ are the same in the deterministic and sampled scenarios.
\par
An alternative to capture stochastic information is representing the distribution by statistical features \cite{Infantes2024}. However, these features are hard to define for complex distributions and may lead to high dimensionality. Moreover, the varied characteristics in the state formulation depend on both the scheduling logic and the processing times, which cannot be accurately described by the statistical features, so valuable information is inevitably lost. 
Our features of the deterministic instance and scenarios can favorably reflect the influence of stochasticity, and are applicable to different problems, feature definitions, and distributions.
Using efficient multiprocessing or parallel GPU computations, the runtime overhead of processing the states of scenarios is negligible, thereby maintaining fast solution construction.

\subsubsection{Action Space}
The action space $\mathcal{A}(t)$ consists of all compatible operation-machine pairs of the first unscheduled operations per job. The same action is applied to the deterministic instance and all the sampled scenarios.

\subsubsection{State Transition}
Upon taking an action, the sets $\mathcal{O}_u(t)$, $\mathcal{M}_u(t)$, and $\mathcal{A}(t)$ are updated to reflect the new relevant operations, machines, and potential actions. Additionally, the features for $s_t^{det}$ and all $s_t^l$ are updated independently, resulting in a new state $s_{t+1}$.

\subsubsection{Reward}
We generalize the reward defined in \cite{zhang2020} and \cite{wang2023flexible}. In this deterministic reward function, a lower bound $\underbar{C}(O_{ij}, s_t)$ of the completion time is estimated for each operation $O_{ij}$ at time step $t$. This bound equals the scheduled completion time if the operation has been scheduled. For unscheduled operations, the lower bound is approximated using the recursion $\underbar{C}(O_{ij}, s_t) = \underbar{C}(O_{i(j-1)}, s_t) + \min_{k \in \mathcal{M}_{ij}} p_{ij}^k$. Then the reward at step $t$ is $r_t = \max_{O_{ij} \in \mathcal{O}}\underbar{C}(O_{ij}, s_t) - \max_{O_{ij} \in \mathcal{O}}\underbar{C}(O_{ij}, s_{t+1})$.
\par
For the stochastic reward, we consider $n_{rew}$ scenarios that are independently sampled and do not overlap with the scenarios of the state, to facilitate the generalization.
These scenarios require a small amount of computation for tracking their makespan lower bounds since they do not necessitate computing all features. 
Concretely, let $\hat{s}_t^l$ represent the state of reward scenario $l$, then $\mathcal{C}_t = \left\{\max_{O_{ij} \in \mathcal{O}}\underbar{C}(O_{ij}, \hat{s}_t^l) | 1 \leq l \leq n_{rew}\right\}$ are the makespan lower bounds of all scenarios at time $t$. The reward is computed as $r_t = f({\mathcal{C}}_t) - f(\mathcal{C}_{t+1})$, where $f$ is the objective function, which in our case is the $VaR_\alpha$ or mean makespan. However, other functions such as conditional $VaR_\alpha$ or median are also possible. If the discounting factor is $\gamma=1$, the sum of all rewards naturally sums to $\sum_{t=0}^{|\mathcal{O}|-1} r_t = f(\mathcal{C}_0) - f(\mathcal{C}_{|\mathcal{O}|-1})$, where $\mathcal{C}_{|\mathcal{O}|-1} = \{C_{max}^l|1 \leq l \leq n_{rew}\}$ are the makespans of all scenarios. Hence, maximizing the cumulative reward corresponds to minimizing the objective.

\subsection{Network Architecture}
\begin{figure*}[t]  
\begin{subfigure}{0.95\columnwidth}
    \centering
    \includegraphics[width=\linewidth]{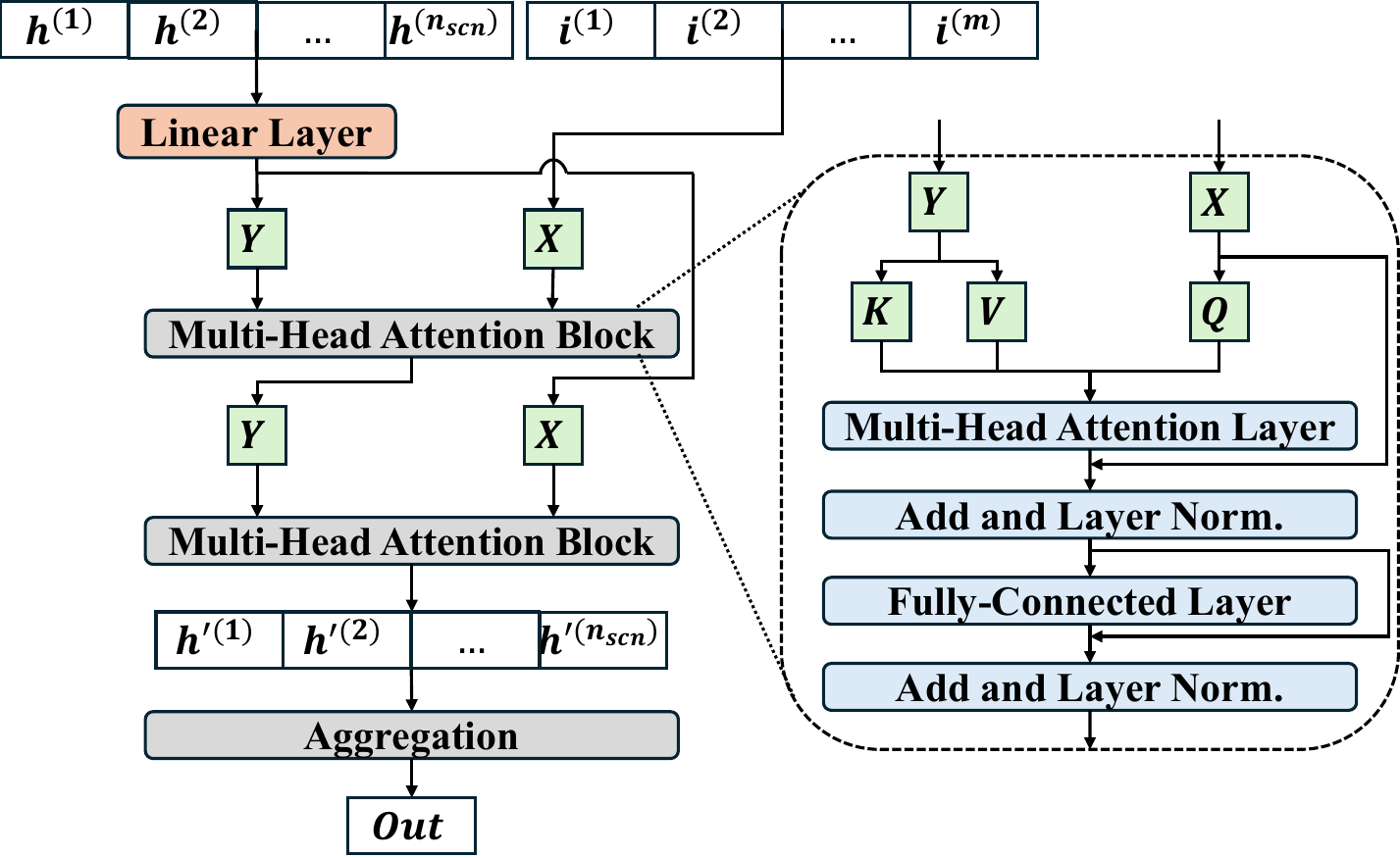}
    \caption{Scenario Processing Module (SPM).}\label{fig:SPM}
\end{subfigure}\hfill
\begin{subfigure}{1.09\columnwidth}
    \centering
    \includegraphics[width=\linewidth]{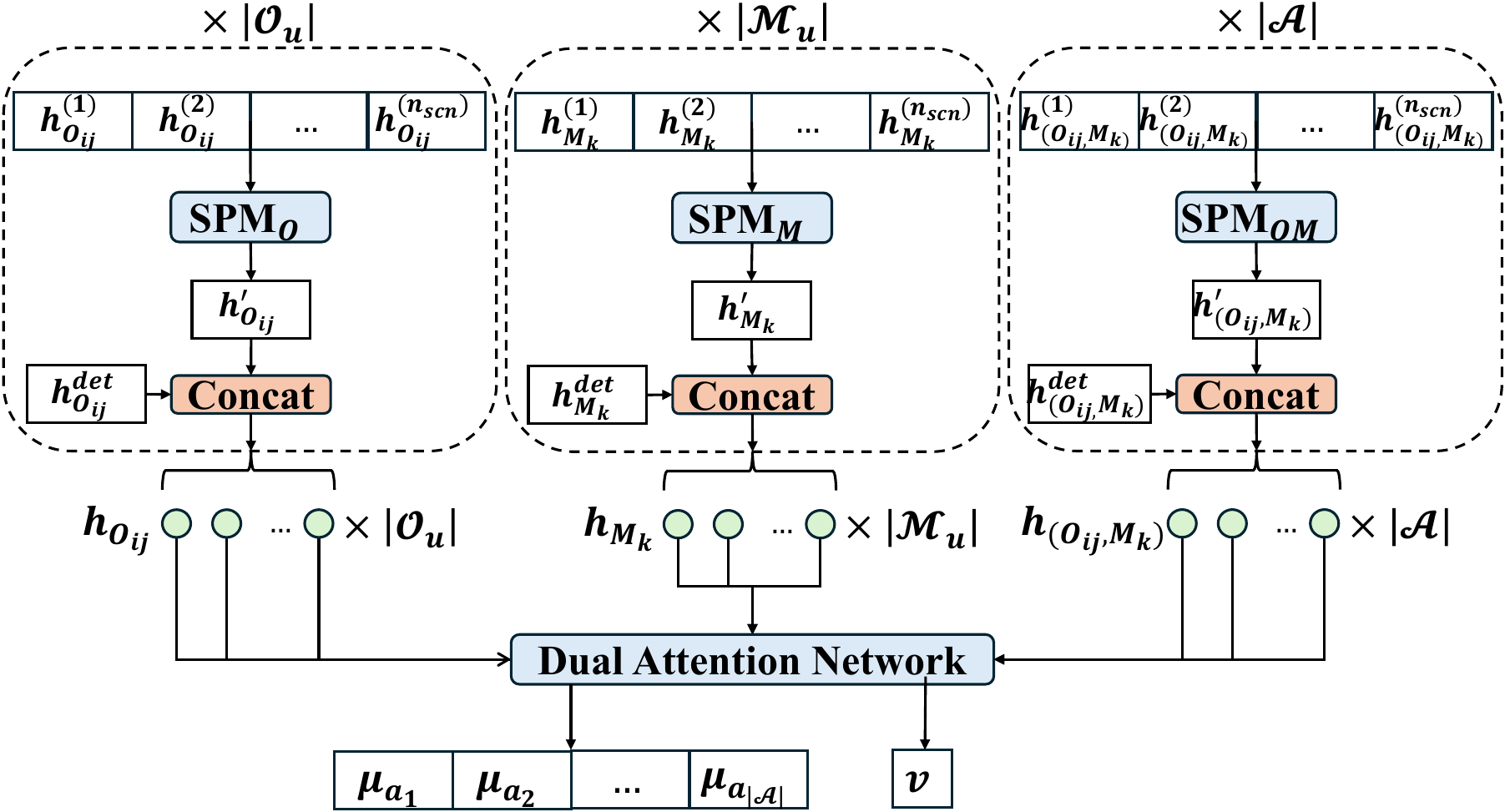}
    \caption{Applying SPM to DAN \cite{wang2023flexible}.}\label{fig:SPM_DAN_link}
\end{subfigure}
\caption{Overview of SPM-DAN.}\label{fig:overview_Network}
\end{figure*}

Our neural network consists of the scenario processing modules (SPMs) that take the scenarios $\mathcal{S}_t^{stoch}$ and compute embeddings from them to represent stochasticity. These are combined with the deterministic state $s_t^{det}$ and inputted to the base neural network. In doing so, we do not need to restructure the base neural network drastically, making our method easily applicable to different GNN architectures. 
An alternative would be to pass all scenarios through the GNN and do `post-processing' before the actor and critic layers. However, this is significantly more computationally complex. 
Figure \ref{fig:overview_Network} outlines the proposed SPM and its application to the state-of-the-art DAN for FJSP \cite{wang2023flexible}.

\subsubsection{Scenario Processing Module}
To efficiently and effectively extract embeddings from the scenario states to reflect the stochasticity, we design the SPM by a multi-head attention mechanism, which is inspired by \cite{Lee2019}. Consider a set $\mathcal{H}$ of $n_{scn}$ scenario states, which are linearly transformed into $d$-dimensional embeddings $H\in \mathbb{R}^{n_{scn} \times d}$. We expect to extract an embedding that represents the distribution of scenario states. To this end, we capture the interaction between scenarios using the multi-head attention. A straightforward way would be to apply the full self-attention $\text{MHAB}(H, H)$. However, self-attention complexity scales quadratically with $n_{scn}$, making it too expensive for a large value $n_{scn}$. Hence, we propose a trainable set of $m$ $d$-dimensional inducing point vectors $I\in \mathbb{R}^{m \times d}$ to compute the cross-attention over $H\in \mathbb{R}^{n_{scn} \times d}$, thereby maintaining a linear complexity with $n_{scn}$ and resulting in the embeddings $J \in \mathbb{R}^{m\times d}$. We then apply the original set $H$ to perform cross-attention with $J$, to get $H'\in \mathbb{R}^{n_{scn}\times d}$, which approximates $\text{MHAB}(H,H)$.  
To extract a single embedding of the stochasticity, we must apply permutation invariant aggregation, for which we use mean aggregation. Formally, our scenario processing module (SPM) can be expressed by:
{
\[\text{SPM}(\mathcal{H}) = \text{Avg}(\text{MHAB}(H, \text{MHAB}(I, H))\]}

\subsubsection{SPM-DAN}
The DAN in \cite{wang2023flexible} comprises $L$ operation message attention blocks (OMBs) and machine message attention blocks (MMBs), followed by the actor and critic network. Each $\text{OMB}_l$ computes embeddings $h_{O_{ij}}^{(l)}$ of operations $O_{ij} \in \mathcal{O}_u$ by an attention mechanism:
{
\[\{h_{O_{ij}}^{(l)} | O_{ij} \in \mathcal{O}_u\} = \text{OMB}_l\left(\{h_{O_{ij}}^{(l-1)} | O_{ij} \in \mathcal{O}_u\}\right)\]}

Similarly, $\text{MMB}_l$ computes embeddings $h_{M_k}^{(l)}$ for all machines $M_k \in \mathcal{M}_u$ through an attention mechanism by involving both the operation and machine features, such that:
{
\begin{equation*}
    \begin{split}
        \{h_{M_k}^{(l)} | M_k \in \mathcal{M}_u\} = \text{MMB}_l\big(& \{h_{M_k}^{(l-1)} | M_k \in \mathcal{M}_u\},\\  &  \{h_{O_{ij}}^{(l-1)} | O_{ij} \in \mathcal{O}_u\}\big) 
    \end{split}
\end{equation*}}

The actor network in the DAN is a multilayer perceptron (MLP) that calculates a score $\mu_{(O_{ij}, M_k)}$ for each action $(O_{ij}, M_k) \in \mathcal{A}$:
{
\[\mu_{(O_{ij}, M_k)} = \text{MLP}_\theta\left([h_{O_{ij}}^{(L)} || h_{M_k}^{(L)} || h_G^{(L)} || h_{(O_{ij}, M_k)}]\right)\]}

\noindent where $h_G^{(L)} = [\frac{1}{|\mathcal{O}_u|} \sum_{O_{ij}\in \mathcal{O}_u}h_{O_{ij}}^{(L)} || \frac{1}{|\mathcal{M}_u|} \sum_{M_{k}\in \mathcal{M}_u}h_{M_{k}}^{(L)}]$ is a graph embedding. Then, the DAN uses the softmax function over the scores to transform them into a probability distribution $\pi_\theta$, from which the action can be sampled. The critic network is an MLP for outputting a scalar value:
{
\[v = \text{MLP}_\phi(h_G^{(L)})\]}

\noindent which estimates the value of actions taken by the actor. We refer to \cite{wang2023flexible} for details of each block.
\par
The DAN requires three distinct types of input features, $h_{O_{ij}} = h_{O_{ij}}^{(0)}$, $h_{M_k} = h_{M_k}^{(0)}$, and $h_{(O_{ij}, M_k)}$. We compute these features based on our state definition. Specifically, we apply our SPM to process the scenario states $\mathcal{S}^{stoch}$ and concatenate the resulting embeddings with their deterministic counterparts, such that:
{\small
\[h_{O_{ij}} = [h_{O_{ij}}^{det} || \text{SPM}_O(\{h_{O_{ij}}^l | 1 \leq l \leq n_{scn} \})]\]
\[h_{M_k} = [h_{M_k}^{det} || \text{SPM}_M(\{h_{M_k}^l | 1 \leq l \leq n_{scn} \})]\]
 \[h_{(O_{ij}, M_k)} = [ h_{(O_{ij}, M_k)}^{det} || \text{SPM}_{OM}( \{h_{(O_{ij}, M_k)}^l | 1 \leq l \leq n_{scn} \})]\]}

In summary, SPM-DAN uses the SPM to compute
embeddings representing the stochasticity of the problem instance. Combined with the deterministic features and passed to DAN, they enable the learning of effective solution policies for solving the stochastic FJSP.

\subsection{Training Procedure}
Following \cite{song2022flexible} and \cite{wang2023flexible}, we train our policy by an actor-critic-based proximal policy optimization (PPO) algorithm \cite{schulman2017}, shown in Algorithm \ref{alg:training-procedure}. We extend the default PPO by keeping track of multiple independent scenarios per problem instance and applying the same actions. We compute the reward and state using these distinct scenarios and combine them into a single transition tuple reflecting all scenarios per problem instance per step. Hence, we maintain many scenarios, accommodating our state space to reflect the stochastic problem, without increasing the PPO batch sizes.
The model is trained for $N_{ep}$ episodes. We sample batches of $n_B$ problem instances, with $n_{scn}$ scenarios to compute the state and $n_{rew}$ scenarios for the reward after $N_s$ episodes. We validate on a fixed set of $n_{vali}$ instances every $N_{eval}$ episodes. As a common practice, we sample actions from the distribution $\pi_\theta$ for training and infer greedy solutions for validation.

\begin{algorithm}[t]
\caption{Training Procedure}\label{alg:training-procedure}
\begin{algorithmic}[1]
\Require {Neural network with randomly initialized parameters, fixed set of $n_{vali}$ evaluation instances}
    \State {Sample batch of $n_B$ instances}
    \For{$n_{ep}=1, 2, \dots N_{ep}$}
    \For{$b=1,2, \dots n_B$} \Comment{In Parallel}
    \For{$t=0,1, \dots, |\mathcal{O}| - 1$}
    \State{Compute $\pi_\theta$ and sample $a_{t,b} \sim \pi_\theta$} 
    \State{Perform $a_{t,b}$ and receive $s_{t+1,b}^{det}$}
    \For{$l=1,2,\dots, n_{scn}$} \Comment{In Parallel}
        \State{Perform $a_{t,b}$ and receive $s_{t+1,b}^l$}
    \EndFor
    \For{$l=1,2, \dots, n_{rew}$} \Comment{In Parallel}
        \State{Perform $a_{t,b}$ and receive $\hat{s}_{t+1,b}^l$}
    \EndFor
    \State{Compute $r_{t,b}$}
    \State{Collect transition $(s_{t,b}, a_{t,b}, r_{t,b}, s_{t+1,b})$}
    \EndFor
    \State{Compute generalized advantage estimates}
    \EndFor
    \State{Compute PPO loss and update network parameters}
    \If{$n_{ep} \text{ mod } N_{eval}=0$}
    \State{Validate the policy}
    \EndIf
    \If{$n_{ep} \text{ mod } N_{B}=0$}
    \State{Sample new batch of $n_B$ training instances}
    \EndIf
    \EndFor
\end{algorithmic}
\end{algorithm}

\section{Experiments}
In this section, we numerically evaluate our method on various problem instances and compare the performance with several baseline methods. 

\subsubsection{Baselines}
We use the first-in-first-out (FIFO), most-operations-remaining (MOR), shortest-processing-time (SPT), and most-work-remaining (MWKR) priority dispatching rules (PDRs). In addition, we evaluate the deterministic OR-tools CP-SAT solver \cite{perron2023} with the median processing times, using the implementation by \cite{reijnen2023}. We also extend the CP-SAT formulation to optimize over multiple sampled scenarios simultaneously, which we name CP-stoch. For these methods, we produce a plan using the deterministic information (or the used scenarios for CP-stoch) and evaluate on $n_{rew}$ independent scenarios. We set a 1-hour time limit for the CP-SAT solver. Based on preliminary tests, we set the number of scenarios for CP-stoch to 25 for synthetic 10$\times$5 and 20$\times$5 instances, and 10 for the other. Lastly, we evaluate the deterministic DAN policies on stochastic instances and use DAN with our reward mechanism but without SPM (DAN-stoch).

\subsubsection{Datasets} 
As is common, we use synthetic datasets for training and evaluation. We use $\text{SD}_1$ from \cite{wang2023flexible}. However, $\text{SD}_1$ has processing times in the range $\{1,\dots,20\}$. Since it is common to round to integers, the stochasticity is limited, especially for small numbers (e.g., 1 with 50\% standard deviation will mostly be 1). We exclude $\text{SD}_2$ from \cite{wang2023flexible} as it has an unrealistic assumption of unrelated processing times $p_{ij}^k \in \{1,\dots,99\}$ for an operation $O_{ij}$, drastically limiting the effects of stochasticity (e.g., $p_{ij}^1=10$ will mostly be favored over $p_{ij}^2=90$ regardless of the variance).
Instead, we propose a more realistic $\text{SD}_3$ for which we mimic the instance structure of $\text{SD}_2$, but sample processing times 
using $\overline{p}_{ij} \sim U(1, 99)$ for each $O_{ij}$, after which we sample $p_{ij}^k \sim U(0.85\overline{p}_{ij}, 1.15\overline{p}_{ij})$. We use instance sizes 10$\times$5, 20$\times$5, 15$\times$10, 20$\times$10, 30$\times$10, and 40$\times$10, with $n\times m$ indicating $n$ jobs and $m$ machines.
\par
We also evaluate our method using the mk \cite{Brandimarte1993}, rdata,  edata, and vdata instances \cite{Hurink1994}. Note, that the mk instances suffer from both limitations of $\text{SD}_1$ and $\text{SD}_2$. 

\subsubsection{Creating Stochastic Instances}
We assume the processing times of the deterministic instances to be the median value. For each pair $(O_{ij}, M_k)$, we sample $CV_{ij}^k\sim U(0.1, 0.5)$ and set the standard deviation $\sigma_{ij}^k = CV_{ij}^k \times median_{ij}^k$, as different machine-operation pairs have various uncertainty levels. There is no agreement on the best probability distributions in the literature. Unless specified otherwise, we assume log-normal distributions like \cite{caldeira2021}. 

\begin{table*}[th]
\small
\setlength{\tabcolsep}{0.51mm}
\renewcommand\arraystretch{0.84}
\begin{tabular}{@{}ccc|cccc|ccc|ccc|cc@{}}
\toprule
                     &       &      & \multicolumn{4}{c|}{PDRs}             & \multicolumn{3}{c|}{Greedy}                    & \multicolumn{3}{c|}{Sample}            &         &          \\
                     & Size  &      & FIFO    & MOR     & MWKR    & SPT     & DAN             & DAN-stoch & SPM-DAN          & DAN     & DAN-stoch & SPM-DAN          & CP-SAT      & CP-stoch \\ \midrule
                     &       & Obj  & 160.66  & 157.40  & 155.01  & 171.16  & \textbf{149.03} & 150.99    & 149.58           & 140.90  & 141.64    & \textbf{140.44}  & 140.60  & 139.07   \\
                     & 10$\times$5  & Gap  & 14.27\% & 11.95\% & 10.25\% & 21.74\% & \textbf{6.00\%} & 7.39\%    & 6.39\%           & 0.21\%  & 0.74\%    & \textbf{-0.11\%} & 0.00\%  & -1.09\%  \\
                     &       & Time (s)& 0.06    & 0.06    & 0.06    & 0.06    & 0.69   & 0.69      & 0.84             & 4.53    & 4.59      & 5.33             & -       & -        \\ \cmidrule(l){2-15} 
                     &       & Obj  & 270.60  & 271.81  & 270.93  & 293.45  & 270.14          & 261.82    & \textbf{255.47}  & 259.75  & 265.59    & \textbf{250.81}  & 251.26  & 260.14   \\
                     & 20$\times$5  & Gap  & 7.70\%  & 8.18\%  & 7.83\%  & 16.79\% & 7.51\%          & 4.20\%    & \textbf{1.68\%}  & 3.38\%  & 5.70\%    & \textbf{-0.18\%} & 0.00\%  & 3.53\%   \\
\multirow{2}{*}{\rotatebox[origin=c]{90}{SD$_1$}} &       & Time (s)& 0.12    & 0.12    & 0.12    & 0.12    & 1.35            & 1.39      & 1.66             & 9.10    & 9.15      & 12.62            & -       & -        \\ \cmidrule(l){2-15} 
                     &       & Obj  & 254.20  & 241.74  & 239.70  & 266.19  & 226.48          & 228.12    & \textbf{220.20}  & 216.17  & 218.61    & \textbf{211.01}  & 216.36  & 224.62   \\
                     & 15$\times$10 & Gap  & 17.49\% & 11.73\% & 10.79\% & 23.03\% & 4.68\%          & 5.44\%    & \textbf{1.77\%}  & -0.09\% & 1.04\%    & \textbf{-2.47\%} & 0.00\%  & 3.82\%   \\
                     &       & Time (s)& 0.19    & 0.19    & 0.19    & 0.18    & 2.10            & 2.07      & 2.45             & 14.41   & 14.42     & 22.50            & -       & -        \\ \cmidrule(l){2-15} 
                     &       & Obj  & 308.87  & 299.23  & 296.27  & 337.68  & 279.44          & 274.36    & \textbf{262.53}  & 270.15  & 266.34    & \textbf{256.77}  & 276.42  & 291.56   \\
                     & 20$\times$10 & Gap  & 11.74\% & 8.25\%  & 7.18\%  & 22.16\% & 1.09\%          & -0.75\%   & \textbf{-5.02\%} & -2.27\% & -3.65\%   & \textbf{-7.11\%} & 0.00\%  & 5.48\%   \\
                     &       & Time (s)& 0.26    & 0.27    & 0.26    & 0.26    & 2.77            & 2.83      & 3.36             & 19.84   & 19.88     & 34.47            & -       & -        \\ \midrule
                     &       & Obj  & 760.14  & 755.55  & 744.38  & 827.24  & 722.77          & 736.05    & \textbf{710.63}  & 688.74  & 693.49    & \textbf{677.24}  & 688.10  & 676.19   \\
                     & 10$\times$5  & Gap  & 10.47\% & 9.80\%  & 8.18\%  & 20.22\% & 5.04\%          & 6.97\%    & \textbf{3.27\%}  & 0.09\%  & 0.78\%    & \textbf{-1.58\%} & 0.00\%  & -1.73\%  \\
                     &       & Time (s)& 0.06    & 0.06    & 0.06    & 0.06    & 0.65            & 0.65      & 0.79             & 0.74    & 0.73      & 1.11             & -       & -        \\ \cmidrule(l){2-15} 
                     &       & Obj  & 1310.67 & 1329.94 & 1320.25 & 1427.00 & 1307.48         & 1290.02   & \textbf{1279.70} & 1272.08 & 1257.08   & \textbf{1243.75} & 1253.67 & 1271.71  \\
                     & 20$\times$5  & Gap  & 4.55\%  & 6.08\%  & 5.31\%  & 13.83\% & 4.29\%          & 2.90\%    & \textbf{2.08\%}  & 1.47\%  & 0.27\%    & \textbf{-0.79\%} & 0.00\%  & 1.44\%   \\
\multirow{2}{*}{\rotatebox[origin=c]{90}{SD$_3$}} &       & Time (s)& 0.12    & 0.12    & 0.12    & 0.12    & 1.28            & 1.37      & 1.58             & 1.46    & 1.48      & 2.99             & -       & -        \\ \cmidrule(l){2-15} 
                     &       & Obj  & 1215.14 & 1183.13 & 1156.25 & 1311.01 & 1111.55         & 1129.99   & \textbf{1090.92} & 1066.60 & 1080.32   & \textbf{1039.73} & 1074.88 & 1105.90  \\
                     & 15$\times$10 & Gap  & 13.05\% & 10.07\% & 7.57\%  & 21.97\% & 3.41\%          & 5.13\%    & \textbf{1.49\%}  & -0.77\% & 0.51\%    & \textbf{-3.27\%} & 0.00\%  & 2.89\%   \\
                     &       & Time (s)& 0.18    & 0.19    & 0.18    & 0.18    & 1.96            & 1.96      & 2.36             & 2.69    & 2.74      & 6.51             & -       & -        \\ \cmidrule(l){2-15} 
                     &       & Obj  & 1449.91 & 1435.71 & 1421.03 & 1640.16 & 1376.18         & 1379.80   & \textbf{1289.77} & 1336.97 & 1346.62   & \textbf{1261.80} & 1351.58 & 1406.68  \\
                     & 20$\times$10 & Gap  & 7.28\%  & 6.22\%  & 5.14\%  & 21.35\% & 1.82\%          & 2.09\%    & \textbf{-4.57\%} & -1.08\% & -0.37\%   & \textbf{-6.64\%} & 0.00\%  & 4.08\%   \\
                     &       & Time (s)& 0.25    & 0.26    & 0.26    & 0.25    & 2.63            & 2.60      & 3.14             & 3.80    & 3.78      & 10.50            & -       & -        \\ \bottomrule
\end{tabular}
\caption{Results on synthetic instances of the same sizes as the instances used for training.}\label{tab:results-synthetic}
\end{table*}

\begin{table*}[ht!]
\small
\setlength{\tabcolsep}{1.mm}
\renewcommand\arraystretch{0.91}
\begin{tabular}{@{}ccc|cccc|ccc|ccc|c@{}}
\toprule
                     &              &      & \multicolumn{4}{c|}{PDRs}             & \multicolumn{3}{c|}{Greedy}            & \multicolumn{3}{c|}{Sample}            &         \\
                     & Size         &      & FIFO    & MOR     & MWKR    & SPT     & DAN     & DAN-stoch & SPM-DAN          & DAN     & DAN-stoch & SPM-DAN          & CP-SAT      \\ \midrule
                     &              & Obj  & 412.05  & 410.85  & 410.41  & 458.12  & 386.09  & 371.58    & \textbf{355.85}  & 376.92  & 365.36    & \textbf{352.16}  & 379.06  \\
                     & 30$\times$10 & Gap  & 8.70\%  & 8.39\%  & 8.27\%  & 20.86\% & 1.85\%  & -1.97\%   & \textbf{-6.12\%} & -0.56\% & -3.61\%   & \textbf{-7.10\%} & 0.00\%  \\
\multirow{2}{*}{\rotatebox[origin=c]{90}{SD$_1$}} &              & Time (s)& 0.40    & 0.41    & 0.41    & 0.40    & 4.16    & 4.15      & 4.93             & 46.80   & 33.40     & 56.67            & -       \\ \cmidrule(l){2-14} 
                     &              & Obj  & 520.59  & 527.47  & 527.98  & 577.27  & 497.46  & 474.88    & \textbf{455.55}  & 488.34  & 469.54    & \textbf{453.01}  & 487.47  \\
                     & 40$\times$10 & Gap  & 6.79\%  & 8.21\%  & 8.31\%  & 18.42\% & 2.05\%  & -2.58\%   & \textbf{-6.55\%} & 0.18\%  & -3.68\%   & \textbf{-7.07\%} & 0.00\%  \\
                     &              & Time (s)& 0.58    & 0.58    & 0.58    & 0.57    & 5.58    & 5.44      & 6.70             & 42.20   & 41.88     & 89.22            & -       \\ \midrule
                     &              & Obj  & 1957.33 & 1997.42 & 1992.04 & 2251.80 & 1917.87 & 1915.07   & \textbf{1776.47} & 1882.99 & 1884.82   & \textbf{1775.37} & 1885.18 \\
                     & 30$\times$10 & Gap  & 3.83\%  & 5.95\%  & 5.67\%  & 19.45\% & 1.73\%  & 1.59\%    & \textbf{-5.77\%} & -0.12\% & -0.02\%   & \textbf{-5.82\%} & 0.00\%  \\
\multirow{2}{*}{\rotatebox[origin=c]{90}{SD$_3$}} &              & Time (s)& 0.41    & 0.41    & 0.41    & 0.40    & 3.95    & 3.95      & 4.76             & 6.17    & 6.40      & 22.26            & -       \\ \cmidrule(l){2-14} 
                     &              & Obj  & 2496.87 & 2559.63 & 2572.53 & 2858.96 & 2474.38 & 2457.34   & \textbf{2294.33} & 2438.07 & 2432.05   & \textbf{2305.52} & 2442.27 \\
                     & 40$\times$10   & Gap  & 2.24\%  & 4.81\%  & 5.33\%  & 17.06\% & 1.31\%  & 0.62\%    & \textbf{-6.06\%} & -0.17\% & -0.42\%   & \textbf{-5.60\%} & 0.00\%  \\
                     &              & Time (s)& 0.57    & 0.57    & 0.57    & 0.56    & 5.20    & 5.26      & 6.29             & 8.64    & 8.63      & 37.64            & -       \\ \bottomrule
\end{tabular}
\caption{Results on large synthetic instances using the policies trained on size 20$\times$10.}\label{tab:results-synthetic-large}
\end{table*}

\subsubsection{Configurations}
For a fair comparison, we keep the same training configurations $N_{ep}=1000$, $N_B=20$, $N_s=20$, $N_{eval}=10$ as \cite{wang2023flexible}. We also keep the same hyperparameters for the OMB and MMB blocks, the critic network, and the PPO algorithm, for which we refer to their paper. We increase the hidden layer size of the actor network to 128 to facilitate the expanded size of the input features $h_{(O_{ij}, M_k)}$. Based on preliminary tests, we use 4 attention heads and dimensions $d=32$ for the SPMs. We set the number of inducing points $m=16$. For training, validation, and testing we set $n_{rew}=1000$ and use $n_{scn}=100$. For the synthetic data, we test 100 problem instances per instance type. We use an NVIDIA A100 GPU and 18-core Intel Xeon Platinum 8360Y CPU for DRL training and a 32-core AMD Rome 7H12 CPU for the CP-SAT solver. We train policies on SD$_1$ and SD$_3$ for the smallest four instance sizes, which takes between 1 and 14 hours. For inference, we use greedy solution construction and a sampling strategy with 100 samples per instance, conforming to the existing FJSP papers. Unless specified otherwise, we assume the $VaR_{95\%}$ of the makespan as the objective. We evaluate this metric using $n_{rew}=1000$ scenarios for each instance. We also report the inference time and the gap to the CP-SAT solutions, calculated as $gap = \frac{f(C_{max}) - f(C_{max})_{CP}}{f(C_{max})_{CP}}$. The reported values are the average for each set of instances.

\begin{table*}[th]
\small
\setlength{\tabcolsep}{0.62mm}
\renewcommand\arraystretch{0.91}
\begin{tabular}{@{}cc|cccc|ccc|ccc|cc@{}}
\toprule
                       &      & \multicolumn{4}{c|}{PDRs}             & \multicolumn{3}{c|}{Greedy}                    & \multicolumn{3}{c|}{Sample}                     &         &          \\
Dataset              &      & FIFO    & MOR     & MWKR    & SPT     & DAN             & DAN-stoch & SPM-DAN          & DAN              & DAN-stoch & SPM-DAN          & CP-SAT      & CP-stoch \\ \midrule
\multirow{3}{*}{mk}    & Obj  & 257.45  & 253.37  & 255.39  & 295.30  & \textbf{232.22} & 238.81    & 232.51           & \textbf{223.61}  & 229.21    & 224.01           & 230.20  & 346.91   \\
                       & Gap  & 11.84\% & 10.07\% & 10.94\% & 28.28\% & \textbf{0.88\%} & 3.74\%    & 1.00\%           & \textbf{-2.86\%} & -0.43\%   & -2.69\%          & 0.00\%  & 50.70\%  \\
                       & Time (s)& 0.18    & 0.18    & 0.18    & 0.18    & 1.98            & 1.94      & 2.37             & 14.89            & 14.96     & 21.71            &     -    &      -    \\ \midrule
\multirow{3}{*}{rdata} & Obj  & 1408.60 & 1397.46 & 1391.50 & 1568.94 & 1370.80         & 1366.73   & \textbf{1345.01} & 1318.70          & 1311.90   & \textbf{1294.62} & 1316.22 & 1549.72  \\
                       & Gap  & 7.02\%  & 6.17\%  & 5.72\%  & 19.20\% & 4.15\%          & 3.84\%    & \textbf{2.19\%}  & 0.19\%           & -0.33\%   & \textbf{-1.64\%} & 0.00\%  & 17.74\%  \\
                       & Time (s)& 0.19    & 0.19    & 0.19    & 0.19    & 2.12            & 2.07      & 2.46             & 16.06            & 16.08     & 23.98            &     -    &      -    \\ \midrule
\multirow{3}{*}{edata} & Obj  & 1586.08 & 1570.56 & 1560.18 & 1695.89 & 1534.73         & 1544.10   & \textbf{1520.25} & 1467.36          & 1483.50   & \textbf{1457.05} & 1423.01 & 1553.96  \\
                       & Gap  & 11.46\% & 10.37\% & 9.64\%  & 19.18\% & 7.85\%          & 8.51\%    & \textbf{6.83\%}  & 3.12\%           & 4.25\%    & \textbf{2.39\%}  & 0.00\%  & 9.20\%  \\
                       & Time (s)& 0.19    & 0.19    & 0.19    & 0.19    & 2.10            & 2.11      & 2.49             & 15.96            & 16.09     & 24.11            &      -   &     -     \\ \midrule
\multirow{3}{*}{vdata} & Obj  & 1295.16 & 1293.13 & 1286.46 & 1444.67 & 1281.67         & 1271.83   & \textbf{1227.05} & 1235.49          & 1225.49   & \textbf{1189.42} & 1300.97 & 1336.54  \\
                       & Gap  & -0.45\% & -0.60\% & -1.12\% & 11.05\% & -1.48\%         & -2.24\%   & \textbf{-5.68\%} & -5.03\%          & -5.80\%   & \textbf{-8.57\%} & 0.00\%  & 2.73\%   \\
                       & Time (s)& 0.20    & 0.20    & 0.19    & 0.19    & 2.06            & 2.05      & 2.52             & 16.19            & 16.11     & 24.27            &     -    &      -    \\ \bottomrule
\end{tabular}
\caption{Results on benchmark instances using policies trained on SD$_3$ instances of size 15$\times$10.} \label{tab:results-bench}
\end{table*}

\begin{table*}[ht!]
\small
\setlength{\tabcolsep}{0.62mm}
\begin{tabular}{@{}cc|cccc|ccc|ccc|cc@{}}
\toprule
             &      & \multicolumn{4}{c|}{PDRs}             & \multicolumn{3}{c|}{Greedy}                    & \multicolumn{3}{c|}{Sample}            &         & \multicolumn{1}{l}{}         \\
Size         &      & FIFO    & MOR     & MWKR    & SPT     & DAN             & DAN-stoch & SPM-DAN          & DAN     & DAN-stoch & SPM-DAN          & CP-SAT      & \multicolumn{1}{l}{CP-stoch} \\ \midrule
             & Obj  & 646.99  & 643.20  & 634.07  & 709.00  & \textbf{616.62} & 621.76    & 619.44           & 595.87  & 596.10    & \textbf{591.32}  & 586.89  & 591.72                       \\
10$\times$5  & Gap  & 10.24\% & 9.59\%  & 8.04\%  & 20.81\% & \textbf{5.07\%} & 5.94\%    & 5.55\%           & 1.53\%  & 1.57\%    & \textbf{0.75\%}  & 0.00\%  & 0.82\%                       \\
             & Time (s)& 0.06    & 0.06    & 0.06    & 0.06    & 0.65            & 0.65      & 0.81             & 0.74    & 0.79      & 1.55             & -       & -                            \\ \midrule
             & Obj  & 1172.19 & 1187.68 & 1181.30 & 1280.42 & 1169.24         & 1149.28   & \textbf{1142.68} & 1146.42 & 1145.32   & \textbf{1130.61} & 1120.84 & 1156.04                      \\
20$\times$5  & Gap  & 4.58\%  & 5.96\%  & 5.39\%  & 14.24\% & 4.32\%          & 2.54\%    & \textbf{1.95\%}  & 2.28\%  & 2.18\%    & \textbf{0.87\%}  & 0.00\%  & 3.14\%                       \\
             & Time (s)& 0.12    & 0.12    & 0.12    & 0.12    & 1.28            & 1.33      & 1.61             & 1.46    & 1.64      & 5.05             & -       & -                            \\ \midrule
             & Obj  & 1074.56 & 1044.78 & 1021.75 & 1168.17 & 983.03          & 986.66    & \textbf{968.87}  & 951.44  & 952.00    & \textbf{931.59}  & 949.46  & 991.76                       \\
15$\times$10 & Gap  & 13.18\% & 10.04\% & 7.61\%  & 23.04\% & 3.54\%          & 3.92\%    & \textbf{2.04\%}  & 0.21\%  & 0.27\%    & \textbf{-1.88\%} & 0.00\%  & 4.46\%                       \\
             & Time (s)& 0.18    & 0.19    & 0.18    & 0.18    & 1.96            & 2.00      & 2.43             & 2.69    & 2.96      & 11.19            & -       & -                            \\ \midrule
             & Obj  & 1302.11 & 1288.43 & 1277.09 & 1485.58 & 1237.78         & 1231.46   & \textbf{1181.44} & 1209.97 & 1203.39   & \textbf{1171.56} & 1213.78 & 1286.98                      \\
20$\times$10 & Gap  & 7.28\%  & 6.15\%  & 5.22\%  & 22.39\% & 1.98\%          & 1.46\%    & \textbf{-2.66\%} & -0.31\% & -0.86\%   & \textbf{-3.48\%} & 0.00\%  & 6.03\%                       \\
             & Time (s)& 0.25    & 0.26    & 0.25    & 0.26    & 2.63            & 2.68      & 3.20             & 3.80    & 4.25      & 18.96            & -       & -                            \\ \bottomrule
\end{tabular}
\caption{Results on synthetic $\text{SD}_3$ instances using the expected makespan objective.} \label{tab:results-mean}
\end{table*}

\subsection{Synthetic Datasets Results}

Table \ref{tab:results-synthetic} presents the performance of our methods using test instances generated from the same distributions as the training instances. These results show that our method consistently outperforms all PDRs. In addition, using sampling inference, our model outperforms the deterministic CP-SAT solver on all instance sets. The CP-stoch formulation is only competitive on the 10$\times$5 instances. In larger instances, the additional computational complexity outweighs the increased model accuracy. DAN and DAN-stoch are competitive with our model only on the smallest SD$_1$ instances. For other instance sets, especially larger ones, our method performs considerably better. In our implementation, SPM-DAN adds runtime of roughly 20\% for greedy inference and 3 times for sampling (3.14 vs. 2.63 and 10.50 vs. 3.80 seconds for SD$_3$ 20$\times$10 per instance), but is still relatively fast and leads to clear performance improvements. For example, greedy inference using SPM-DAN considerably outperforms sampling inference using the baseline models for 20$\times$10 instances, showcasing a better performance-runtime trade-off despite an initial runtime increase. The DAN-stoch models do not consistently outperform DAN, showcasing the value of synergizing our training procedure and SPM.
\par
Table \ref{tab:results-synthetic-large} shows the generalization of our policies to larger instances, using the 20$\times$10 policies on 30$\times$10 and 40$\times$10 instances. We do not report CP-stoch, as memory limits are reached before finding sensible solutions for these instance sizes. In these larger instances, our method consistently and considerably outperforms all baselines, even achieving a 6\% improvement over CP-SAT and DAN using greedy sampling. DAN-stoch also beats DAN in these instances but remains far from SPM-DAN performance. These results show that SPM-DAN can scale to new instance sizes without the need for retraining.

\subsection{Public Datasets Results} 
We assess the cross-distribution performance of our method
on the public benchmark datasets, shown in Table \ref{tab:results-bench}. We report the policy trained on 15$\times$10 SD$_3$ instances, but policies trained on 10$\times$5 SD$_3$, 15$\times$10 SD$_1$, and 10$\times$5 SD$_1$ instances show similar results.
We find that our method maintains performance across the benchmark instances. On the mk data, we see no performance improvement of our approach over the default DAN, caused by the previously mentioned instance limitations. On the other datasets, we maintain a clear performance improvement over the baseline neural methods and outperform CP-SAT up to 8\% on vdata instances.

\subsection{Flexibility to Different Objectives} \label{sec:expected_makespan}
We train and evaluate policies for optimizing the expected makespan to show that our method works for different objective functions. Table \ref{tab:results-mean} shows the results of these policies for SD$_3$ instances. We see a similar pattern as for the $VaR_{95\%}$ where there is a limited effect for the smallest instances but our policies outperform the other methods toward larger problem sizes. The overall improvement is slightly less compared to the $VaR_{95\%}$ objective. This is expected as the expected value and deterministic values tend to be more related to each other than the $VaR_{95\%}$ and deterministic values as the $VaR_{95\%}$ is more dependent on the distributional range of processing times. However, the improvements of our method are still considerable and for the larger instances we improve over CP-SAT, as well as the other baselines.

\begin{table*}[th]
\small
\setlength{\tabcolsep}{0.39mm}
\renewcommand\arraystretch{0.91}
\begin{tabular}{@{}cc|cccc|ccc|ccc|cc@{}}
\toprule
                      &      & \multicolumn{4}{c|}{PDRs}             & \multicolumn{3}{c|}{Greedy}            & \multicolumn{3}{c|}{Sample}            &         & \multicolumn{1}{l}{}         \\
Distribution         &      & FIFO    & MOR     & MWKR    & SPT     & DAN     & DAN-stoch & SPM-DAN          & DAN     & DAN-stoch & SPM-DAN          & CP-SAT      & \multicolumn{1}{l}{CP-stoch} \\ \midrule
                      & Obj  & 1267.59 & 1251.52 & 1236.12 & 1452.37 & 1186.82 & 1187.25   & \textbf{1161.30} & 1156.79 & 1159.41   & \textbf{1133.56} & 1170.30 & 1233.08                      \\
B                 & Gap  & 8.31\%  & 6.94\%  & 5.62\%  & 24.10\% & 1.41\%  & 1.45\%    & \textbf{-0.77\%} & -1.15\% & -0.93\%   & \textbf{-3.14\%} & 0.00\%  & 5.36\%                       \\
                      & Time (s) & 0.25    & 0.26    & 0.26    & 0.25    & 2.70    & 2.66      & 3.40             & 5.84    & 5.85      & 18.88            & -       & -                            \\ \midrule
                      & Obj  & 1371.43 & 1356.26 & 1343.65 & 1559.23 & 1298.64 & 1303.83   & \textbf{1229.54} & 1254.58 & 1265.35   & \textbf{1194.14} & 1275.49 & 1322.15                      \\
LB     & Gap  & 7.52\%  & 6.33\%  & 5.34\%  & 22.25\% & 1.81\%  & 2.22\%    & \textbf{-3.60\%} & -1.64\% & -0.79\%   & \textbf{-6.38\%} & 0.00\%  & 3.66\%                       \\
                      & Time (s)& 0.25    & 0.26    & 0.26    & 0.25    & 2.77    & 2.70      & 3.25             & 5.84    & 5.81      & 19.17            & -       & -                            \\ \midrule
                      & Obj  & 1359.61 & 1344.19 & 1329.91 & 1546.49 & 1289.39 & 1278.59   & \textbf{1228.24} & 1247.24 & 1242.22   & \textbf{1197.75} & 1263.08 & 1321.16                      \\
LBG & Gap  & 7.64\%  & 6.42\%  & 5.29\%  & 22.44\% & 2.08\%  & 1.23\%    & \textbf{-2.76\%} & -1.25\% & -1.65\%   & \textbf{-5.17\%} & 0.00\%  & 4.60\%                       \\
                      & Time (s)& 0.25    & 0.26    & 0.26    & 0.25    & 2.74    & 2.66      & 3.25             & 5.50    & 5.87      & 19.17            & -       & -                            \\
                      \bottomrule
\end{tabular}
\caption{Results on synthetic $\text{SD}_3$ instances of size 20$\times$10 using different processing time distributions.} \label{tab:results-distributions}
\end{table*}

\begin{table*}[th]
\small
\setlength{\tabcolsep}{0.4mm}
\renewcommand\arraystretch{0.91}
\begin{tabular}{@{}cc|cccccc|cccccc|cc@{}}
\toprule
             &      & \multicolumn{6}{c|}{Greedy}                                                 & \multicolumn{6}{c|}{Sample}                                                          &         &          \\
             &      & \multicolumn{3}{c}{SPM-DAN}      & \multicolumn{3}{c|}{SPM-DAN-200}     & \multicolumn{3}{c}{SPM-DAN}      & \multicolumn{3}{c|}{SPM-DAN-200}              &         &          \\
Size         &      & 50      & 100     & 200              & 50      & 100     & 200              & 50      & 100     & 200              & 50      & 100              & 200              & CP-SAT      & CP-stoch \\ \midrule
             & Obj  & 719.14  & 710.63  & \textbf{709.58}  & 713.64  & 709.69  & 712.11           & 674.51  & 677.24  & 672.29           & 673.36  & \textbf{671.75}  & 675.07           & 688.10  & 676.19   \\
10$\times$5  & Gap  & 4.51\%  & 3.27\%  & \textbf{3.12\%}  & 3.71\%  & 3.14\%  & 3.49\%           & -1.98\% & -1.58\% & -2.30\%          & -2.14\% & \textbf{-2.38\%} & -1.89\%          & 0.00\%  & -1.73\%  \\
             & Time (s)& 0.80    & 0.79    & 0.81             & 0.80    & 0.81    & 0.78             & 1.10    & 1.11    & 2.64             & 1.12    & 1.55             & 1.52             & -       & -        \\ \midrule
             & Obj  & 1283.09 & 1279.70 & \textbf{1277.45} & 1312.16 & 1307.93 & 1297.82          & 1242.07 & 1243.75 & \textbf{1238.65} & 1268.96 & 1265.04          & 1266.04          & 1253.67 & 1271.71  \\
20$\times$5  & Gap  & 2.35\%  & 2.08\%  & \textbf{1.90\%}  & 4.67\%  & 4.33\%  & 3.52\%           & -0.93\% & -0.79\% & \textbf{-1.20\%} & 1.22\%  & 0.91\%           & 0.99\%           & 0.00\%  & 1.44\%   \\
             & Time (s)& 1.65    & 1.58    & 1.61             & 1.66    & 1.62    & 1.57             & 3.20    & 2.99    & 8.72             & 3.19    & 5.03             & 4.87             & -       & -        \\ \midrule
             & Obj  & 1084.99 & 1090.92 & \textbf{1077.59} & 1092.26 & 1088.52 & 1089.67          & 1035.24 & 1039.73 & \textbf{1035.22} & 1041.44 & 1040.63          & 1044.55          & 1074.88 & 1105.90  \\
15$\times$10 & Gap  & 0.94\%  & 1.49\%  & \textbf{0.25\%}  & 1.62\%  & 1.27\%  & 1.38\%           & -3.69\% & -3.27\% & \textbf{-3.69\%} & -3.11\% & -3.19\%          & -2.82\%          & 0.00\%  & 2.89\%   \\
             & Time (s)& 2.41    & 2.36    & 2.47             & 2.43    & 2.47    & 2.37             & 6.64    & 6.51    & 16.50            & 6.65    & 11.22            & 11.17            & -       & -        \\ \midrule
             & Obj  & 1290.38 & 1289.77 & 1285.83          & 1296.15 & 1295.71 & \textbf{1285.53} & 1261.14 & 1261.80 & \textbf{1256.07} & 1262.78 & 1259.25          & 1259.41          & 1351.58 & 1406.68  \\
20$\times$10 & Gap  & -4.53\% & -4.57\% & -4.86\%          & -4.10\% & -4.13\% & \textbf{-4.89\%} & -6.69\% & -6.64\% & \textbf{-7.07\%} & -6.57\% & -6.83\%          & -6.82\%          & 0.00\%  & 4.08\%   \\
             & Time (s) & 3.24    & 3.14    & 3.33             & 3.24    & 3.30    & 3.17             & 11.10   & 10.50   & 28.22            & 11.09   & 19.09            & 18.61            & -       & -        \\ \midrule
             & Obj  & 1783.31 & 1776.47 & 1776.10          & 1788.87 & 1778.37 & \textbf{1775.00} & 1775.52 & 1775.37 & 1768.42          & 1765.85 & 1759.35          & \textbf{1754.91} & 1885.18 & -        \\
30$\times$10 & Gap  & -5.40\% & -5.77\% & -5.79\%          & -5.11\% & -5.67\% & \textbf{-5.84\%} & -5.82\% & -5.82\% & -6.19\%          & -6.33\% & -6.67\%          & \textbf{-6.91\%} & 0.00\%  & -        \\
             & Time (s)& 4.89    & 4.76    & 5.22             & 4.98    & 4.86    & 4.95             & 23.40   & 22.26   & 62.53            & 23.40   & 33.93            & 61.56            & -       & -        \\ \midrule
             & Obj  & 2297.15 & 2294.33 & 2289.39          & 2303.53 & 2292.60 & \textbf{2286.71} & 2309.34 & 2305.52 & 2299.30          & 2283.10 & 2276.22          & \textbf{2269.88} & 2442.27 & -        \\
40$\times$10 & Gap  & -5.94\% & -6.06\% & -6.26\%          & -5.68\% & -6.13\% & \textbf{-6.37\%} & -5.44\% & -5.60\% & -5.85\%          & -6.52\% & -6.80\%          & \textbf{-7.06\%} & 0.00\%  & -        \\
             & Time (s) & 6.58    & 6.29    & 7.00             & 6.66    & 6.67    & 7.18             & 40.06   & 37.64   & 107.75           & 40.06   & 58.94            & 107.20           & -       & -        \\ \bottomrule
\end{tabular}
\caption{Results on synthetic $\text{SD}_3$ instances of models trained with different $n_{scn}$ using various numbers of $n_{scn}$ for inference.} \label{tab:results-inference-scenarios}
\end{table*}

\subsection{Resilience to Varying Probability Distributions}
An important criterion for stochastic optimization methods is that they perform under different assumed probability distributions. Therefore, we create one instance set with beta distributions (B) similar to \cite{Flores2023}, one set with a mixture of log-normal and beta distributions (LB), and one set with a mixture of log-normal, beta, and gamma distributions (LBG). The results are shown in Table \ref{tab:results-distributions}. It is clear that our method maintains performance across all instance sets. We see that DAN-stoch struggles to outperform DAN, as it cannot capture the dynamics that it observes through the rewards. Oppositely, our SPM-DAN model is aware of the types of stochasticity, which helps it distinguish different stochastic states. As a result, SPM-DAN achieves a gap compared to CP-SAT of 6\% and improves performance up to 5\% over the other models.

\subsection{Trade-Off Performance and Number of Scenarios}
We assess how $n_{scn}$ affects the performance of SPM-DAN. To this end, we evaluate models trained with 100 (SPM-DAN) and 200 (SPM-DAN-200) scenarios using 50, 100, and 200 scenarios for inference. Table \ref{tab:results-inference-scenarios} shows that, at the cost of higher (lower) inference times, using a higher (lower) $n_{scn}$ to generate solutions generally improves (decreases) performance. The differences in performance gap are roughly between 0.5\% and 1\%.  However, training using more scenarios does not per se lead to better performance, especially for smaller instances. Namely, SPM-DAN with 200 evaluation scenarios outperforms SPM-DAN-200 in some instance sets. In larger instances, the performance difference between SPM-DAN and SPM-DAN-200 is limited, although SPM-DAN-200 performs slightly better. Hence, computational requirements can be decreased by training with fewer scenarios at the cost of a limited potential performance loss. Then, in inference, $n_{scn}$ can be varied to trade off runtime and performance.
The increased runtime of using a higher $n_{scn}$ is also limited by increasing parallelism.

\section{Conclusion}
This study proposes the scenario processing module SPM, a novel attention-based module to extend NCO methods to stochastic problems. SPM captures embeddings from sampled scenarios and feeds those to the base neural network, allowing it to learn effective policies under uncertainty. We integrate SPM into a training procedure that works well with different stochastic objectives and apply it to the stochastic FJSP, leading to SPM-DAN. We experimentally show that SPM-DAN outperforms existing learning and non-learning methods. SPM-DAN works well across a range of synthetic and benchmark instances while handling various processing time distributions. Future work may focus on developing and adapting sampling techniques for stochastic optimization to enhance the performance of constructive policies. Moreover, integrating data from historical schedules with their realizations or contextual factors influencing processing times is a worthwhile avenue for exploration.

\section*{Acknowledgements}
The LEO (Learning and Explaining Optimization) project is co-funded by Holland High Tech | TKI HSTM via the PPS allowance scheme for public-private partnerships. This work used the Dutch national e-infrastructure with the support of the SURF Cooperative using grant no. EINF-10518.
\bibliography{aaai25}

\clearpage
\appendix
\setcounter{secnumdepth}{1}
\section{Transferability to Different Base Neural Networks} \label{sec:base-networks}
We provide preliminary results of applying SPM to the L2D method \cite{zhang2020} for the JSP in Table \ref{tab:results-l2d}. The L2D method consists of a completely different network architecture than the DAN. Moreover, the JSP is a structurally different problem than the FJSP with a different graph structure and the L2D uses a different set of features than those used by the DAN \cite{wang2023flexible} for the FJSP. Hence, through this experiment, we can show the strong applicability of our method to different networks and problems. 

We train L2D models for 10x10 instances and evaluate them on different sizes using greedy inference. We use all the default settings from \cite{zhang2020} for the L2D network architecture and PPO algorithm. Since L2D only has one node type with features, we only require one SPM. For the SPM, we use 4 attention heads, set dimension $d=16$, and set the number of inducing points to $m=8$. SPM-L2D outperforms L2D and L2D-stoch for each instance size, even with minimal tuning, showcasing the transferability of our module to different base neural networks and CO problems.

\section{CP-stoch Formulation}\label{sec:cp-stoch}
Algorithm \ref{alg:CP-stoch} on the next page shows how to formulate the CP-stoch model for the CP-SAT solver. The formulation comprises three parts. Firstly, we develop the basic FJSP model for each scenario. Secondly, we enforce that the schedule (i.e., operation-machine assignment and ordering) is the same for each scenario. Thirdly, we define the objective based on the makespans of the scenarios.

\begin{algorithm*}
\small
    \caption{CP-stoch Formulation}\label{alg:CP-stoch}
    \begin{algorithmic}
    \Require{Maximum possible scheduling value Horizon, number of scenarios $n$, set of jobs $\mathcal{J}$, set of machines $\mathcal{M}$, sets of possible machines per operation $\mathcal{M}_{ij}$ processing times per scenario $p_{ij}^{k,l}$, stochastic objective function $f$.}
        \For{$l=1, 2, \dots, n$} \Comment{Create the basic FJSP formulation for each scenario.}
            \For{$i=1,2,\dots, |\mathcal{J}|$}
                \For{$j = 1,2,\dots |\mathcal{O}_{i}|$}
                    \State $p_{ij}^{min,l}=\min_{M_k \in \mathcal{M}_{ij}}p_{ij}^{k,l}$
                    \State $p_{ij}^{max,l}=\max_{M_k \in \mathcal{M}_{ij}}p_{ij}^{k,l}$
                    \State $start_{ij}^l = \texttt{NewIntVar}$(0, Horizon)
                    \State $end_{ij}^l = \texttt{NewIntVar}$(0, Horizon)
                    \State $duration_{ij}^l = \texttt{NewIntVar}$($p_{ij}^{min,l}$, $p_{ij}^{max,l}$)
                    \State $Interval_{ij}^l = \texttt{NewIntervalVar}(start_{ij}^l, duration_{ij}^l, end_{ij}^l)$
                    \State $\texttt{Add}(start_{ij}^l \geq end_{i,j-1}^l)$
                    \For{$M_k \in \mathcal{M}_{ij}$}
                        \State $start_{ij}^{k,l} = \texttt{NewIntVar}$(0, Horizon)
                        \State $end_{ij}^{k,l} = \texttt{NewIntVar}$(0, Horizon)
                        \State $Interval_{ij}^{k,l} = \texttt{NewIntervalVar}(start_{ij}^{k,l}, p_{ij}^{k,l} , end_{ij}^{k,l})$
                        \State $Presence_{ij}^{k,l} = \texttt{NewBoolVar}()$
                        \State $\texttt{Add}(start_{ij}^l = start_{ij}^{k,l}).\texttt{OnlyEnforceIf}(Presence_{ij}^{k,l})$
                        \State $\texttt{Add}(end_{ij}^l = end_{ij}^{k,l}).\texttt{OnlyEnforceIf}(Presence_{ij}^{k,l})$
                        \State $\texttt{Add}(duration_{ij}^l = p_{ij}^{k,l}).\texttt{OnlyEnforceIf}(Presence_{ij}^{k,l})$
                    \EndFor
                    \State $\texttt{AddExactlyOne}(\{Presence_{ij}^{k,l}|\forall M_k \in \mathcal{M}_{ij}\})$
                \EndFor
            \EndFor
            \State $c_{max}^l = \texttt{NewIntVar}$(0, Horizon)
            \State $\texttt{AddMaxEquality}(c_{max}^l,end_{ij}^l)$ 

            \For{$M_k \in \mathcal{M}$}
                \State $\texttt{AddNoOverlap}(\{Interval_{ij}^{k,l}|\forall i=1,2, \dots, |\mathcal{J}|, j = 1,2,\dots |\mathcal{O}_{i}|\})$
            \EndFor
        \EndFor

        \For{$l=1, 2, \dots, n$} \Comment{Ensure that the same schedule is used for each scenario.}
            \For{$M_k \in \mathcal{M}$}
                \For{$Interval_{ij}^{k,l}, Interval_{i'j'}^{k,l} \quad \forall i, i'=1,2,\dots, |\mathcal{J}|, j, j' = 1,2,\dots |\mathcal{O}_{i}|, i' \geq i, j'\geq j$}
                    \State $Interval_{ij}^{k,l}\_before\_Interval_{i'j'}^{k,l} = \texttt{NewBoolVar}()$
                    \State $Interval_{i'j'}^{k,l}\_before\_Interval_{ij}^{k,l} = \texttt{NewBoolVar}()$
                    \State $\texttt{Add}(end_{ij}^{k,l} \leq  start_{i'j'}^{k,l}).\texttt{OnlyEnforceIf}(Interval_{ij}^{k,l}\_before\_Interval_{i'j'}^{k,l})$ 
                    \State $\texttt{Add}(end_{i'j'}^{k,l} \leq  start_{ij}^{k,l}).\texttt{OnlyEnforceIf}(Interval_{i'j'}^{k,l}\_before\_Interval_{ij}^{k,l})$ 
                    \State $\texttt{AddBoolXOr}(Interval_{ij}^{k,l}\_before\_Interval_{i'j'}^{k,l}, Interval_{i'j'}^{k,l}\_before\_Interval_{ij}^{k,l})$
                \EndFor
            \EndFor
        \EndFor
        \For{$M_k \in \mathcal{M}$}
            \For{$Interval_{ij}^{k,l}, Interval_{i'j'}^{k,l} \quad \forall i, i'=1,2,\dots, |\mathcal{J}|, j, j' = 1,2,\dots |\mathcal{O}_{i}|$}
                \State $\texttt{AddAllowedAssignments}(\{Interval_{ij}^{k,l}\_before\_Interval_{i'j'}^{k,l} | l=1, 2, \dots, n \}, ((1,\dots, 1), (0, \dots, 0)))$
            \EndFor
            \For{$i=1,2, \dots, |\mathcal{J}|, j = 1,2,\dots |\mathcal{O}_{i}|$}
               \State $\texttt{AddAllowedAssignments}(\{Presence_{ij}^{k,l} | l=1, 2, \dots, n\}, ((1,\dots, 1), (0, \dots, 0)))$
            \EndFor
        \EndFor
        \If{$f =  Avg$} \Comment{Create objective value.}
            \State{$obj = \texttt{Sum}(\{c_{max}^l | l=1,2,\dots,n\}) / n$}
        \EndIf
        \If{$f = VaR_{95\%}$}
            \For{$l=1, 2, \dots, n$}
                \For{$l'=l+1, l+2, \dots, n$}
                    \State $l\_less\_than\_l' = \texttt{NewBoolVar}()$
                    \State $l'\_less\_than\_l = \texttt{NewBoolVar}()$
                    \State $\texttt{Add}(c_{max}^l \leq c_{max}^{l'}).\texttt{OnlyEnforceIf}(l\_less\_than\_l')$
                    \State $\texttt{Add}(c_{max}^{l'} < c_{max}^{l}).\texttt{OnlyEnforceIf}(l'\_less\_than\_l)$
                    \State $\texttt{AddBoolXOr}(l\_less\_than\_l', l'\_less\_than\_l)$
                \EndFor
            \EndFor
            \For{$l=1, 2, \dots, n$}
                \State $is\_obj^l = \texttt{NewBoolVar}()$
                \State $\texttt{Add}(\texttt{Sum}(\{l\_less\_than\_l'| l'=1,2,\dots,n, l'\neq l\}) = \texttt{round}(0.95n)).\texttt{OnlyEnforceIf}(is\_obj^l)$
            \EndFor
            \State $\texttt{AddBoolXOr}(\{is\_obj^l | l=1, 2, \dots, n\})$
            \State $obj = \texttt{Sum}(\{is\_obj^l \times c_{max}^l | l=1, 2, \dots, n\}$
        \EndIf
        \State $\texttt{Minimize}(obj)$
    \end{algorithmic}
\end{algorithm*}

\section{Public Dataset Results} \label{sec:public_datasets_other_policies}
Tables \ref{tab:results-bench-SD3-10x5}, \ref{tab:results-bench-SD1-15x10}, and \ref{tab:results-bench-SD1-10x5} (on the second page below) show the results of applying policies trained on SD$_3$ 10$\times$5, SD$_1$ 15$\times$10, and SD$_1$ 10$\times$5 policies on the public benchmark datasets. These results are similar to those shown in the main results, which indicates that our method is generally better than the other baselines. It again suggests that the proposed SPM is effective in handling the stochasticity.

\section{Significance of Results} \label{sec:t-tests}
To assess the significance of the improvement of our method, we perform $t$-tests between SPM-DAN and the best alternative between DAN and DAN-stoch. In Tables \ref{tab:t-test-1} and \ref{tab:t-test-2} we show the values of these $t$-tests corresponding to Tables \ref{tab:results-synthetic} and \ref{tab:results-bench}. These results show that the improvement of our model is generally significant. The only instance sets that do not show a significant difference are the smallest $SD_1$ instances and the mk dataset, in which we also remarked very similar performance from the original results. Hence, our model shows significant performance improvement in most instance sets and is never significantly worse than other methods.

\begin{table}[t]
\centering
\begin{tabular}{@{}c|ccc@{}}
\toprule
Size  &      L2D     & L2D-stoch & SPM-L2D   \\ \midrule
10x10 &  1289.09 & 1278.45   & \textbf{1261.49}   \\ 
15x15 &  1964.52 & 1933.45   & \textbf{1907.75}   \\ 
20x10 &  1851.22 & 1842.17        & \textbf{1829.35} \\ \bottomrule
\end{tabular}
\caption{$VaR_{95\%}$ makespan using L2D for the JSP.}\label{tab:results-l2d}
\end{table}

\begin{table}[t]
\centering
\renewcommand\arraystretch{0.85}
\small
\begin{tabular}{@{}ccc|cc@{}}
\toprule 
                     &             Size           &     & Greedy          & Sample          \\ \midrule
\multirow{8}{*}{\rotatebox[origin=c]{90}{SD$_1$}} & \multirow{2}{*}{10x5}  & $t$ & -0.85           & 1.64            \\
                     &                        & $p$ & 0.40            & 0.10            \\ \cmidrule(l){2-5} 
                     & \multirow{2}{*}{20x5}  & $t$ & 9.71            & 27.54           \\
                     &                        & $p$ & \textless{}0.01 & \textless{}0.01 \\ \cmidrule(l){2-5} 
                     & \multirow{2}{*}{15x10} & $t$ & 8.24            & 14.84           \\
                     &                        & $p$ & \textless{}0.01 & \textless{}0.01 \\ \cmidrule(l){2-5} 
                     & \multirow{2}{*}{20x10} & $t$ & 22.25           & 29.99           \\
                     &                        & $p$ & \textless{}0.01 & \textless{}0.01 \\ \midrule
\multirow{8}{*}{\rotatebox[origin=c]{90}{SD$_3$}} & \multirow{2}{*}{10x5}  & $t$ & 3.13            & 6.27            \\
                     &                        & $p$ & \textless{}0.01 & \textless{}0.01 \\ \cmidrule(l){2-5} 
                     & \multirow{2}{*}{20x5}  & $t$ & 3.48            & 9.56            \\
                     &                        & $p$ & \textless{}0.01 & \textless{}0.01 \\ \cmidrule(l){2-5} 
                     & \multirow{2}{*}{15x10} & $t$ & 5.23            & 17.67           \\
                     &                        & $p$ & \textless{}0.01 & \textless{}0.01 \\ \cmidrule(l){2-5} 
                     & \multirow{2}{*}{20x10} & $t$ & 25.07           & 44.15           \\
                     &                        & $p$ & \textless{}0.01 & \textless{}0.01 \\ \bottomrule
\end{tabular}
\caption{The $t$-test of SPM-DAN compared with the best alternative between DAN and DAN-stoch, with the results of Table \ref{tab:results-synthetic}.} \label{tab:t-test-1}
\end{table}

\begin{table}[t]
\centering
\renewcommand\arraystretch{0.85}
\small
\begin{tabular}{@{}cc|cc@{}}
\toprule
Dataset                &     & Greedy          & Sample          \\ \midrule
\multirow{2}{*}{mk}    & $t$ & -0.09           & -0.52           \\
                       & $p$ & 0.93            & 0.62            \\ \midrule
\multirow{2}{*}{rdata} & $t$ & 3.47            & 7.65            \\
                       & $p$ & \textless{}0.01 & \textless{}0.01 \\ \midrule
\multirow{2}{*}{edata} & $t$ & 1.90            & 2.57            \\
                       & $p$ & 0.07            & 0.01            \\ \midrule
\multirow{2}{*}{vdata} & $t$ & 6.35            & 8.73            \\
                       & $p$ & \textless{}0.01 & \textless{}0.01 \\ \bottomrule
\end{tabular}
\caption{The $t$-tests of SPM-DAN compared with the best alternative between DAN and DAN-stoch, with the results of Table \ref{tab:results-bench}.} \label{tab:t-test-2}
\end{table}

\begin{table*}[th]
\small
\setlength{\tabcolsep}{0.57mm}
\begin{tabular}{@{}cc|cccc|ccc|ccc|cc@{}}
\toprule
                       &      & \multicolumn{4}{c|}{PDRs}             & \multicolumn{3}{c|}{Greedy}                           & \multicolumn{3}{c|}{Sample}            &         &          \\
Benchmark              &      & FIFO    & MOR     & MWKR    & SPT     & DAN             & DAN-stoch        & SPM-DAN          & DAN     & DAN-stoch & SPM-DAN          & CP-SAT      & CP-stoch \\ \midrule
\multirow{3}{*}{mk}    & Obj  & 257.45  & 253.37  & 255.39  & 295.30  & \textbf{238.92} & 245.51           & 240.61           & 230.60  & 235.61    & \textbf{229.11}  & 230.20  & 346.91   \\
                       & Gap  & 11.84\% & 10.07\% & 10.94\% & 28.28\% & \textbf{3.79\%} & 6.65\%           & 4.52\%           & 0.17\%  & 2.35\%    & \textbf{-0.47\%} & 0.00\%  & 50.70\%  \\
                       & Time & 0.18    & 0.18    & 0.18    & 0.18    & 1.96            & 1.95             & 2.32             & 14.84   & 14.83     & 21.70            & -       & -        \\ \midrule
\multirow{3}{*}{rdata} & Obj  & 1408.60 & 1397.46 & 1391.50 & 1568.94 & 1390.54         & 1401.33          & \textbf{1363.55} & 1329.26 & 1322.77   & \textbf{1310.99} & 1316.22 & 1549.72  \\
                       & Gap  & 7.02\%  & 6.17\%  & 5.72\%  & 19.20\% & 5.65\%          & 6.47\%           & \textbf{3.60\%}  & 0.99\%  & 0.50\%    & \textbf{-0.40\%} & 0.00\%  & 17.74\%  \\
                       & Time & 0.19    & 0.19    & 0.19    & 0.19    & 2.09            & 2.07             & 2.49             & 15.97   & 16.11     & 24.06            & -       & -        \\ \midrule
\multirow{3}{*}{edata} & Obj  & 1586.08 & 1570.56 & 1560.18 & 1695.89 & 1574.54         & \textbf{1527.72} & 1536.81          & 1480.75 & 1461.58   & \textbf{1458.65} & 1423.01 & 1553.96  \\
                       & Gap  & 11.46\% & 10.37\% & 9.64\%  & 19.18\% & 10.65\%         & \textbf{7.36\%}  & 8.00\%           & 4.06\%  & 2.71\%    & \textbf{2.50\%}  & 0.00\%  & 9.20\%   \\
                       & Time & 0.19    & 0.19    & 0.19    & 0.19    & 2.08            & 2.13             & 2.48             & 16.00   & 15.91     & 24.29            & -       & -        \\ \midrule
\multirow{3}{*}{vdata} & Obj  & 1295.16 & 1293.13 & 1286.46 & 1444.67 & 1299.53         & 1297.38          & \textbf{1239.23} & 1244.04 & 1241.20   & \textbf{1197.39} & 1300.97 & 1336.54  \\
                       & Gap  & -0.45\% & -0.60\% & -1.12\% & 11.05\% & -0.11\%         & -0.28\%          & \textbf{-4.75\%} & -4.38\% & -4.59\%   & \textbf{-7.96\%} & 0.00\%  & 2.73\%   \\
                       & Time & 0.20    & 0.20    & 0.19    & 0.19    & 2.07            & 2.05             & 2.52             & 16.15   & 16.03     & 24.00            & -       & -        \\ \bottomrule
\end{tabular}
\caption{Results on benchmark instances using policies trained on SD$_3$ instances of size 10$\times$5.} \label{tab:results-bench-SD3-10x5}
\end{table*}

\begin{table*}[th]
\small
\setlength{\tabcolsep}{0.57mm}
\begin{tabular}{@{}cc|cccc|ccc|ccc|cc@{}}
\toprule
                       &      & \multicolumn{4}{c|}{PDRs}             & \multicolumn{3}{c|}{Greedy}            & \multicolumn{3}{c|}{Sample}            &         &          \\
Benchmark              &      & FIFO    & MOR     & MWKR    & SPT     & DAN     & DAN-stoch & SPM-DAN          & DAN     & DAN-stoch & SPM-DAN          & CP-SAT      & CP-stoch \\ \midrule
\multirow{3}{*}{mk}    & Obj  & 257.45  & 253.37  & 255.39  & 295.30  & 235.11  & 238.81    & \textbf{232.80}  & 226.31  & 229.21    & \textbf{222.91}  & 230.20  & 346.91   \\
                       & Gap  & 11.84\% & 10.07\% & 10.94\% & 28.28\% & 2.13\%  & 3.74\%    & \textbf{1.13\%}  & -1.69\% & -0.43\%   & \textbf{-3.17\%} & 0.00\%  & 50.70\%  \\
                       & Time & 0.18    & 0.18    & 0.18    & 0.18    & 1.99    & 1.94      & 2.34             & 14.81   & 14.96     & 21.64            & -       & -        \\ \midrule
\multirow{3}{*}{rdata} & Obj  & 1408.60 & 1397.46 & 1391.50 & 1568.94 & 1381.59 & 1366.73   & \textbf{1344.37} & 1324.28 & 1311.90   & \textbf{1293.61} & 1316.22 & 1549.72  \\
                       & Gap  & 7.02\%  & 6.17\%  & 5.72\%  & 19.20\% & 4.97\%  & 3.84\%    & \textbf{2.14\%}  & 0.61\%  & -0.33\%   & \textbf{-1.72\%} & 0.00\%  & 17.74\%  \\
                       & Time & 0.19    & 0.19    & 0.19    & 0.19    & 2.10    & 2.07      & 2.49             & 16.16   & 16.08     & 24.06            & -       & -        \\ \midrule
\multirow{3}{*}{edata} & Obj  & 1586.08 & 1570.56 & 1560.18 & 1695.89 & 1562.53 & 1544.10   & \textbf{1529.24} & 1485.20 & 1483.50   & \textbf{1473.65} & 1423.01 & 1553.96  \\
                       & Gap  & 11.46\% & 10.37\% & 9.64\%  & 19.18\% & 9.80\%  & 8.51\%    & \textbf{7.47\%}  & 4.37\%  & 4.25\%    & \textbf{3.56\%}  & 0.00\%  & 9.20\%   \\
                       & Time & 0.19    & 0.19    & 0.19    & 0.19    & 2.08    & 2.11      & 2.51             & 15.93   & 16.09     & 24.15            & -       & -        \\ \midrule
\multirow{3}{*}{vdata} & Obj  & 1295.16 & 1293.13 & 1286.46 & 1444.67 & 1286.41 & 1271.83   & \textbf{1214.20} & 1240.34 & 1225.49   & \textbf{1183.97} & 1300.97 & 1336.54  \\
                       & Gap  & -0.45\% & -0.60\% & -1.12\% & 11.05\% & -1.12\% & -2.24\%   & \textbf{-6.67\%} & -4.66\% & -5.80\%   & \textbf{-8.99\%} & 0.00\%  & 2.73\%   \\
                       & Time & 0.20    & 0.20    & 0.19    & 0.19    & 2.08    & 2.05      & 2.51             & 16.12   & 16.11     & 23.98            & -       & -        \\ \bottomrule
\end{tabular}
\caption{Results on benchmark instances using policies trained on SD$_1$ instances of size 15$\times$10.} \label{tab:results-bench-SD1-15x10}
\end{table*}

\begin{table*}[th]
\small
\setlength{\tabcolsep}{0.59mm}
\begin{tabular}{@{}cc|cccc|ccc|ccc|cc@{}}
\toprule
                       &      & \multicolumn{4}{c|}{PDRs}             & \multicolumn{3}{c|}{Greedy}            & \multicolumn{3}{c|}{Sample}                   &         &          \\
Benchmark              &      & FIFO    & MOR     & MWKR    & SPT     & DAN     & DAN-stoch & SPM-DAN          & DAN     & DAN-stoch        & SPM-DAN          & CP-SAT      & CP-stoch \\ \midrule
\multirow{3}{*}{mk}    & Obj  & 257.45  & 253.37  & 255.39  & 295.30  & 238.12  & 239.42    & \textbf{236.22}  & 229.01  & 231.21           & \textbf{226.72}  & 230.20  & 346.91   \\
                       & Gap  & 11.84\% & 10.07\% & 10.94\% & 28.28\% & 3.44\%  & 4.01\%    & \textbf{2.62\%}  & -0.52\% & 0.44\%           & \textbf{-1.51\%} & 0.00\%  & 50.70\%  \\
                       & Time & 0.18    & 0.18    & 0.18    & 0.18    & 1.95    & 1.97      & 2.37             & 14.92   & 15.00            & 21.74            & -       & -        \\ \midrule
\multirow{3}{*}{rdata} & Obj  & 1408.60 & 1397.46 & 1391.50 & 1568.94 & 1375.27 & 1365.25   & \textbf{1361.59} & 1323.63 & \textbf{1309.42} & 1311.20          & 1316.22 & 1549.72  \\
                       & Gap  & 7.02\%  & 6.17\%  & 5.72\%  & 19.20\% & 4.49\%  & 3.73\%    & \textbf{3.45\%}  & 0.56\%  & \textbf{-0.52\%} & -0.38\%          & 0.00\%  & 17.74\%  \\
                       & Time & 0.19    & 0.19    & 0.19    & 0.19    & 2.10    & 2.07      & 2.46             & 16.13   & 16.09            & 24.02            & -       & -        \\ \midrule
\multirow{3}{*}{edata} & Obj  & 1586.08 & 1570.56 & 1560.18 & 1695.89 & 1551.14 & 1530.58   & \textbf{1529.12} & 1484.88 & 1467.90          & \textbf{1466.16} & 1423.01 & 1553.96  \\
                       & Gap  & 11.46\% & 10.37\% & 9.64\%  & 19.18\% & 9.00\%  & 7.56\%    & \textbf{7.46\%}  & 4.35\%  & 3.15\%           & \textbf{3.03\%}  & 0.00\%  & 9.20\%   \\
                       & Time & 0.19    & 0.19    & 0.19    & 0.19    & 2.06    & 2.09      & 2.55             & 15.86   & 16.00            & 24.11            & -       & -        \\ \midrule
\multirow{3}{*}{vdata} & Obj  & 1295.16 & 1293.13 & 1286.46 & 1444.67 & 1288.45 & 1277.37   & \textbf{1270.95} & 1240.82 & 1230.14          & \textbf{1223.41} & 1300.97 & 1336.54  \\
                       & Gap  & -0.45\% & -0.60\% & -1.12\% & 11.05\% & -0.96\% & -1.81\%   & \textbf{-2.31\%} & -4.62\% & -5.44\%          & \textbf{-5.96\%} & 0.00\%  & 2.73\%   \\
                       & Time & 0.20    & 0.20    & 0.19    & 0.19    & 2.04    & 2.04      & 2.51             & 16.13   & 16.09            & 24.03            & -       & -        \\ \bottomrule
\end{tabular}
\caption{Results on benchmark instances using policies trained on SD$_1$ instances of size 10$\times$5.} \label{tab:results-bench-SD1-10x5}
\end{table*}

\end{document}